\documentclass[conference]{IEEEtran}
\IEEEoverridecommandlockouts
\usepackage{cite}
\usepackage{amsmath,amssymb,amsfonts}
\usepackage{algorithmic}
\usepackage{graphicx}
\usepackage{textcomp}
\usepackage{xcolor}
\usepackage{subfig}
\def\BibTeX{{\rm B\kern-.05em{\sc i\kern-.025em b}\kern-.08em
    T\kern-.1667em\lower.7ex\hbox{E}\kern-.125emX}}

\begin{document}

\title{Linear Matrix Factorization Embeddings for Single-objective Optimization Landscapes\\
\thanks{Our work was financially supported by the Slovenian Research Agency (research core funding No. P2-0098 and project No. Z2-1867) and by the Paris Ile-de-France region. We also acknowledge support by COST Action CA15140 ``Improving Applicability of Nature-Inspired Optimisation by Joining Theory and Practice (ImAppNIO)''. }
}

\makeatletter
\newcommand{\linebreakand}{%
  \end{@IEEEauthorhalign}
  \hfill\mbox{}\par
  \mbox{}\hfill\begin{@IEEEauthorhalign}
}
\makeatother

\author{\IEEEauthorblockN{1\textsuperscript{st} Tome Eftimov}
\IEEEauthorblockA{\textit{Computer Systems Department} \\
\textit{Jo\v{z}ef Stefan Institute}\\
Ljubljana, Slovenia \\
tome.eftimov@ijs.si}
\and
\IEEEauthorblockN{2\textsuperscript{nd} Gorjan Popovski}
\IEEEauthorblockA{\textit{Computer Systems Department} \\
\textit{Jo\v{z}ef Stefan Institute}\\
Ljubljana, Slovenia \\
gorjan.popovski@ijs.si}
\and
\IEEEauthorblockN{3\textsuperscript{rd} Quentin Renau}
\IEEEauthorblockA{\textit{Thales and \'Ecole Polytechnique} \\
\textit{Institut Polytechnique de Paris}\\
Palaiseau, France \\
quentin.renau@thalesgroup.com}
 \linebreakand % <------------- \and with a line-break
\IEEEauthorblockN{4\textsuperscript{th} Peter Koro\v{s}ec}
\IEEEauthorblockA{\textit{Computer Systems Department} \\
\textit{Jo\v{z}ef Stefan Institute}\\
Ljubljana, Slovenia \\
peter.korosec@ijs.si}
\and
\IEEEauthorblockN{5\textsuperscript{th} Carola Doerr}
\IEEEauthorblockA{\textit{LIP6} \\
\textit{Sorbonne Université, CNRS}\\
Paris, France \\
carola.doerr@lip6.fr}
}

\maketitle

\begin{abstract}
Automated per-instance algorithm selection and configuration have shown promising performances for a number of classic optimization problems, including satisfiability, AI planning, and TSP. The techniques often rely on a set of features that measure some characteristics of the problem instance at hand. In the context of black-box optimization, these features have to be derived from a set of $(x,f(x))$ samples. A number of different features have been proposed in the literature, measuring, for example, the modality, the separability, or the ruggedness of the instance at hand. Several of the commonly used features, however, are highly correlated. While state-of-the-art machine learning techniques can routinely filter such correlations, they hinder explainability of the derived algorithm design techniques. 

We therefore propose in this work to pre-process the measured (raw) landscape features through representation learning. More precisely, we show that a linear dimensionality reduction via matrix factorization significantly contributes towards a better detection of correlation between different problem instances -- a key prerequisite for successful automated algorithm design. 
\end{abstract}


\section{Introduction}
Supervised machine learning techniques can support the user in selecting a best suitable optimization heuristics for a given problem instance by providing automated data-driven recommendations~\cite{HutterKV19, kerschke_automated_2019}. 
Such machine-trained approaches require meaningful features that quantify different aspects of the problem instance at hand. 

We are concerned in this work with \emph{black-box optimization,} where the problem instances are not accessible other than by evaluating the quality of solution candidates (e.g., through experiments or computer simulations). In this context, the features (apart from generic ones that only describe the decision space itself but not its relationship with the objective values) have to be derived from sampling this decision space. Extracting meaningful features from these samples is studied under the notion of \emph{exploratory landscape analysis}~\cite{mersmann_exploratory_2011}. Especially for the case of numerical black-box optimization, a large number of features has been suggested over the last decades~\cite{DerbelLVAT19,flacco2019}. Each of these features measures a different characteristic of the problem instance. Building upon these features, several works have demonstrated the feasibility of automated algorithm selection~\cite{kerschke_automated_2019,MunozSurvey15} and configuration~\cite{BelkhirDSS17} in the black-box scenario. However, it is well known that several of the existing feature sets contain highly correlated features. In terms of performance, this poses problems only to na\"ive learning techniques, since state-of-the-art ML can handle such correlations in an automated way. However, when the focus is less on achieving peak performance but rather on \emph{understanding} the underlying recommendations, then these correlations pose a problem, as they hinder the explainability of the automated recommendations. It is therefore common practice to invest substantial computing resources in a proper feature selection~\cite{KerschkeT19}.  

We propose in this work an alternative way to deal with redundant information in the feature sets. We demonstrate that representation learning via matrix factorization (singular value decomposition (SVD)~\cite{golub1971singular}) can effectively detect correlated features and can properly normalize their importance via so-called fingerprint embeddings. With this decomposition, new feature vectors can easily be mapped into comparable \emph{fingerprints} of much higher expressiveness than the original feature value. 

We demonstrate the effectiveness of the representation learning approach by applying it to detect correlations between different instances obtained from the 24 functions of the BBOB benchmark set (black-box optimization benchmark~\cite{cocoplat}), a standard benchmark set of numerical optimization problems. While it is difficult to detect which instances are obtained from the same problem when using the original feature vectors (either normalized or non-normalized features values), we obtain high accuracy when computing instance correlation based on the fingerprint representations. We also perform a stratified 5-fold cross  classification of the instances, and observe that classifiers trained with the embeddings are more robust than those trained with the original features values.

\textbf{Availability of Code and Data:} The code and data of this project are available at~\cite{data}, feature values at~\cite{Quentin}.

\section{Background}
\label{sec:background}

We provide some basic background on 
%the core concepts of our study. These are 
Exploratory Landscape Analysis %(Sec.~\ref{sec:ELA}) 
and Representation Learning.% (Sec.~\ref{sec:RL}). 

%%%%%%%%%%%%%%%%%%%%%%%%%%%%%%%%%
\subsection{Exploratory Landscape Analysis}
\label{sec:ELA}

Exploratory landscape analysis (ELA) was introduced in~\cite{mersmann_exploratory_2011} to support the user of black-box optimization heuristics through machine-trained recommendations for the algorithm design that best suits the problem at hand. ELA builds on fitness landscape analysis~\cite{MalanE13surveyFLA,Stadler02} and translates it into a setting in which the underlying problem is only available as a query-able black-box and not -- as required in classical fitness landscape analysis -- as an explicit model $f:S \rightarrow \mathbb{R}$. The key idea of ELA is to describe the characteristics of an optimization problem through a set of features, where each feature measures a different aspect of the problem. As the problems are black-box, the feature values have to be estimated from an (ideally small) set of samples, where each sample is an $(x,f(x))$ pair, which requires evaluation of the solution candidate $x$.    
A number of ELA features can be conveniently computed by the R-package \emph{flacco}~\cite{flacco2019}, which comprises up to 343 feature values, which are grouped into 17 feature sets~\cite{kerschke_automated_2019}. 

% %%%%%%%%%%%%%%%%%%%%%%%%%%%%%%%%%
% \subsection{Benchmark Problems}
% \label{sec:benchmarks}

%%%%%%%%%%%%%%%%%%%%%%%%%%%%%%%%%
\subsection{Representation Learning}
\label{sec:RL}
 Representation learning aims at mapping data into new embeddings which store the essential information of the original data set in a more appropriate and often more dense form, and this via automated mappings~\cite{bengio2013representation}. The representations can catch, for example, redundancies in the data, or effects that stem from a misfortunate scaling of the data. They are also successfully used to reduce the dimension of the data, via automatically detecting correlations. Representation learning has its most important applications in machine learning, where bias and redundancies in data can have severe effects on performance.  
 
 Among the most popular representation learning approaches are unsupervised techniques, which require only limited human domain knowledge and interaction. Matrix factorization is one of these unsupervised techniques. It has proven to be effective in a broad range of applications, see~\cite{RLFeatureSelection,wang2020structured,wang2015subspace} for examples in feature selection.  

\section{Methodology}
\label{sec:methodology}
To describe the matrix factorization approach in our context of ELA applications, let us assume that we have $n$ benchmark problems, for each of which we consider $m$ different instances. 
Let us further assume that for each instance we calculate $k$ features via the ELA sampling approach described above. We organize the obtained data in an $nm \times k$ matrix $X$ with entries that we refer to as $X_{i,j;t}$. That is, for $i \in [1..n]$, $j \in [1..m]$, and $t\in [1..k]$, the value $X_{i,j;t}$ corresponds to the $t$-th feature value obtained for the $j$-th instance of the $i$-th benchmark problem, and the row 
$\mathbf{p_{i,j}}^{T}=\left[X_{i,j;1}, X_{i,j;2},\cdots, X_{i,j;k}\right]$ is the vector of feature values for this instance. 

As mentioned above, several standard ELA features are highly correlated, which may bias the interpretation of the feature vectors, by giving stronger weight to highly redundant information. To address this issue, we propose to learn a vector representation (i.e., an \emph{embedding}) for the feature vectors by applying \emph{singular value decomposition} (SVD)~\cite{golub1971singular}. That is, we decompose the matrix $X$ into 
$X=U\Sigma V^T$, where $U$ and $V$ are are orthogonal matrices, $\Sigma$ is a diagonal matrix with singular values on the diagonal, and 
the column vectors of $U$ and $V$ correspond to the left and right singular vectors, respectively. The 
benchmark problem instance embedding can be conveniently calculated as: $\hat{\mathbf{p}}_\mathbf{{i,j}} = \Sigma^{-1}V^T\mathbf{p_{i,j}}.$ 

By applying SVD, we map (``embed'') our instance vectors $\mathbf{p_{i,j}}^{T}$ to another vector space in which the new vectors are presented in different uncorrelated dimensions of the data. 
The space of the embedded vectors is defined by the singular value components, where each instance embedding is a combination of the contribution of each singular value. Note also that SVD performs a \textit{linear dimensionality reduction}. When the matrix $X$ is a square matrix, the learned embeddings are unit-norm vectors, whereas the norm of the learned embeddings are less than one otherwise. This follows from the fact that the column vectors of $U$ are always unit-norm vectors. 

In most cases, the matrix $X$ is a full-rank matrix so that, necessarily, the embedded vector space has the same dimension as the matrix. However, a low-rank approximation $X_r$ for the matrix $X$ can also be calculated by this approach. To this end, the $r$ largest singular values and only the first $r$ columns of $U$ and $V$ are used to express 
$X_r=U_r\Sigma_r {V_r}^T$. 
The choice of $r$ is basically a trade-off between time and accuracy, as smaller $r$ can save computing time at the risk of an increased approximation accuracy error. We will analyse the impact of the size of $r$ on our ELA data in Section~\ref{sec:BBOB}. 
Approaches to estimate the number of useful components are the \emph{non-graphical Cattell’s Scree test}~\cite{raiche2013non} and so-called \emph{parallel analysis}~\cite{hayton2004factor}.  

\section{Experiments}
\label{sec:experiments}

Our test-bed are the 24 benchmark problems from the BBOB benchmark set~\cite{bbob-functions} of the COCO (Comparing Continuous Optimizers) platform~\cite{cocoplat}. 
For each problem, we consider the first five instances of the functions in 5D. These instances all stem from the same base problem, and are obtained by translations in search and objective space. The BBOB benchmark set is a well established environment to benchmark derivative-free numerical optimization techniques and forms the basis for discussions in the BBOB workshop series at the annual ACM GECCO conference. 

From the \textit{flacco} software mentioned in Section~\ref{sec:ELA} we use 38 features, which stem from the following five sets: dispersion (disp), information content (ic), nearest better clustering (nbc), meta model (ela\_meta), and $y$-distribution (ela\_distr). 
These 38 features have been chosen because they do not require adaptive sampling, nor have they been dismissed as inappropriate in previous works. 
For example, it has been shown that some of the \emph{flacco} features exhibit low robustness to the random  sampling~\cite{renau_expressiveness_2019,GallagherECJ19TrueValue} and to instance transformations such as translation and scaling~\cite{vskvorc2020understanding}. 

To address a potential lack of feature value robustness, we use median values that are computed from 100 independent approximations. Each approximation is based on 1250 samples, which are generated using the Python package \emph{sobol\_seq} (version 0.1.2), 
%\quentin{Link is \url{https://github.com/naught101/sobol_seq} and version is 0.1.2}, 
with randomly chosen initial seeds.
This package computes so-called Sobol' sequences~\cite{sobol_distribution_1967}, a construction that is known to provide samples of small star discrepancy~\cite{Mat99}. 
We note that different sampling strategies are known to provide different feature values~\cite{Renau-sensitive}. We have therefore repeated some of the analyses presented in this work with uniformly chosen samples and obtained results that are comparable to those reported in the subsequent sections. We omit a detailed discussion of these results for clarity of presentation.

\subsection{Homogeneity of the BBOB Instances}
\label{sec:BBOB}

To explore the homogeneity of the BBOB instances, we first study the Pearson correlation of the vector representations of the 38 (non-normalized) features mentioned in Section~\ref{sec:ELA}. 
The Pearson correlation was chosen since SVD performs linear dimensionality reduction and the Pearson correlation is a measure for the linear correlation between two vectors. 
We plot these pairwise correlations of the $5 \cdot 24$ problem instances in a heatmap, which can be found in Figure~\ref{fig:BBOB_feature_space_cor}. This heatmap lists all BBOB instances in the sequence of the problems, i.e., we start with instance 1 of function $f_1$ on the lower left corner, which is followed by instance 2 of function 1 to the right and to the top, and we continue this way up to instance 5 of problem $f_{24}$ in the rightmost column and in the uppermost row. 
 We clearly observe that almost all instances show high correlation, either positive or negative. Only three exceptions to this picture exist, and those are the second instance of $f_{14}$, the first instance of $f_{17}$, and the fourth instance of $f_{19}$. Their feature vectors show no linear correlation with the feature vectors of the other problem instances, and this is not the case even for the other instances of the same problem.  
%  Using the figure most of the instances are highly correlated, no matter to which benchmark problem they belong. There are only three instances that are not linearly correlated with the other, i.e., the second instance of $f_{14}$, the first instance of $f_{17}$, and the fourth instance of $f_{19}$. 
If asked to interpret this data, one would likely be tempted to conclude %This result indicates 
that there is no reason to select and use most of the instances, as they hardly show sufficient complementarity. % because they do not contribute to a diverse benchmark problem set. 
Even when replacing the original feature values by normalized values, the result does not change much in term of outcomes from the correlation analysis, because the normalization alone does only reduce direction not the intensity of the linear correlation. Using the normalized original features, for most of the instances, we are still not able to distinguish between instances from the same problem.
%Similar result is also obtained if we used normalized features values. By normalizing them, we change the feature space, which results only in changing correlations direction and intensity.

\begin{figure*}[t]
 \subfloat[Correlations between BBOB instances using their representations with 38 landscape features in the original space]{\label{fig:BBOB_feature_space_cor}
      \includegraphics[width=.47\textwidth]{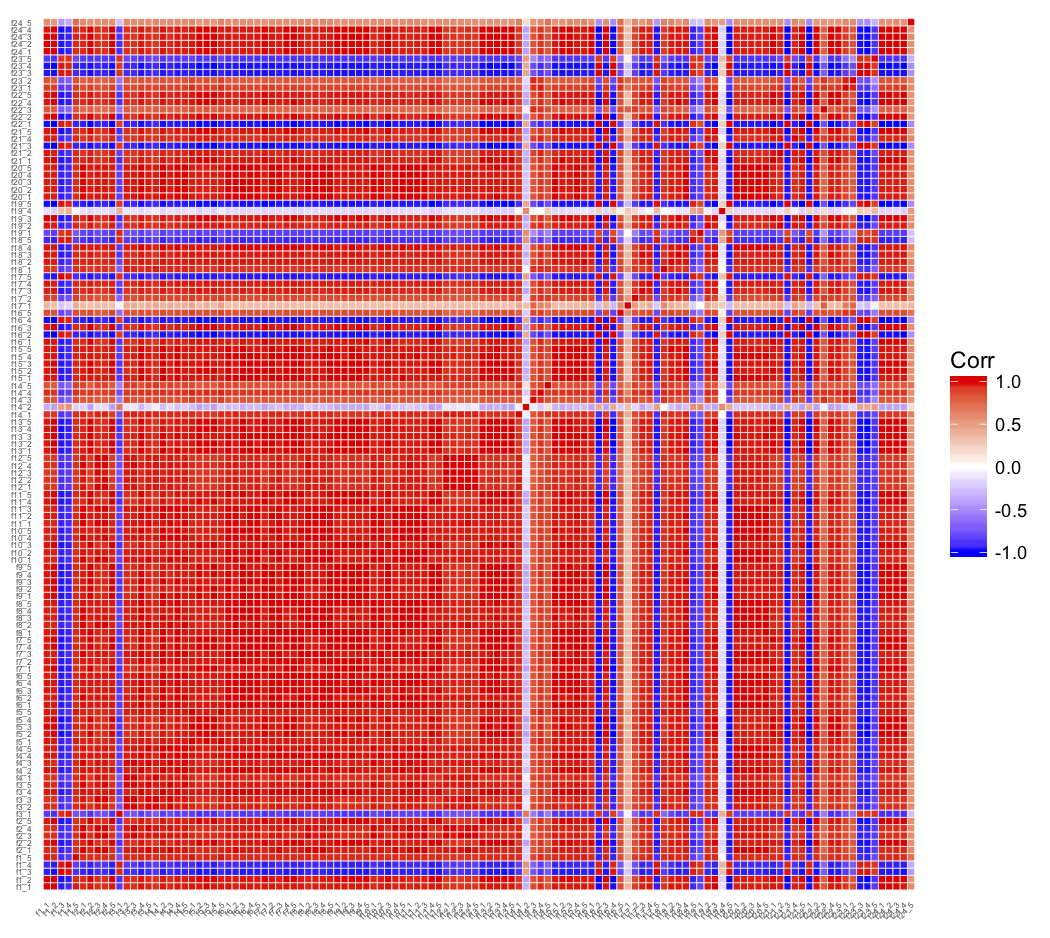}}
~
   \subfloat[Correlations between landscape features using their representations, which are the values obtained for each BBOB instance.]{\label{fig:features_corr}
      \includegraphics[width=.47\textwidth]{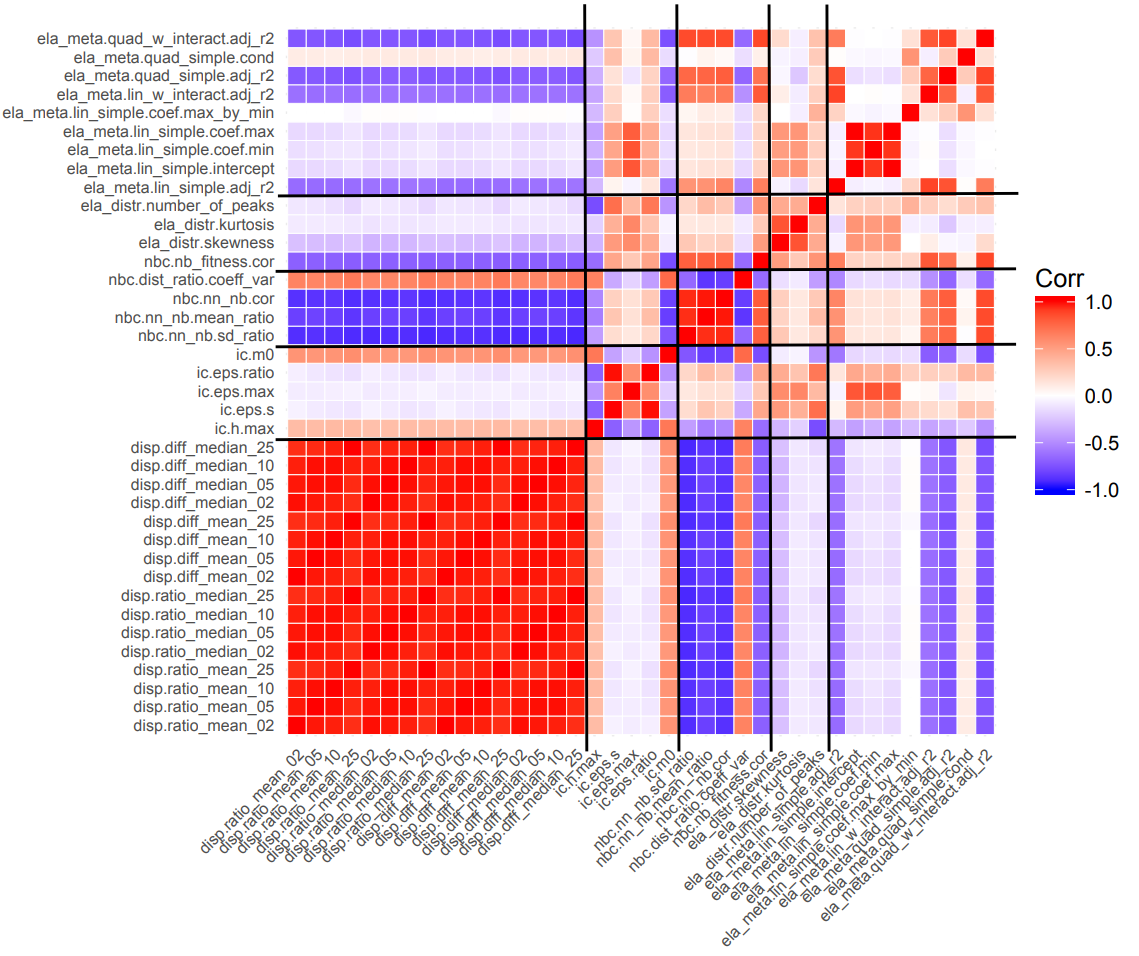}}

   \caption{Pearson correlation analysis. }\label{fig:BBOB_original_space}
\end{figure*}

%\begin{figure}
  %  \centering
  %  \includegraphics[scale=0.7]{figures/FeatureSpace_1250.pdf}
   % \caption{Pearson correlation between BBOB instances using their representations of 38 landscape features.}
    %\label{fig:BBOB_feature_space_cor}
%\end{figure}

A simple analysis helps to explain the effects that lead to the high pairwise correlation of almost all 120 tested BBOB functions. We simply transpose the feature matrix and compute the Pearson correlation between the landscape features -- as opposed to computing correlation between the instances. 
Figure~\ref{fig:features_corr} shows the result of this experiment, and we clearly observe a very high correlation between the features belonging to the same feature set. For example, the dispersion features on the lower left corner show high positive correlation between all feature pairs. Since the sizes of the feature sets are not identical, more weight is given to the larger ones when computing the correlation between the instances. This explains, to a large extent, the high correlation observed in Figure~\ref{fig:BBOB_feature_space_cor}, as we shall see in the next paragraphs.  
% presents the Pearson correlation between the landscape features, using their values obtained for each BBOB instance. It is obvious that the features from the same feature group are highly correlated. Additionally, the unequal number of features from each group can also affect the correlations between the instances. 
%In most applications, the features are selected by subject-matter experts.
%\begin{figure}
   % \centering
 %   \includegraphics[scale=0.45]{figures/features_1250_png.png}
  %  \caption{Pearson correlation between landscape features using their representations, which are the values obtained for each BBOB instance.}
  %  \label{fig:features_corr}
%\end{figure}

The high correlation between the feature values observed in Figure~\ref{fig:features_corr} indicates a strong need for representation learning. We apply the suggested matrix factorization approach described in Section~\ref{sec:methodology}. That is, we first compute the embeddings (``fingerprints'') for each benchmark problem instance. Since, we are interested in high accuracy (i.e., low approximation error from the decomposition) and since computational efficiency is not a problem when dealing with such (comparatively) small matrix sizes ($120 \times 38$),  % size of a matrix involved in our experiment, 
we have used all 38 singular values to calculate the embeddings.

\begin{figure}[tb]
    \centering
    \includegraphics[scale=0.45]{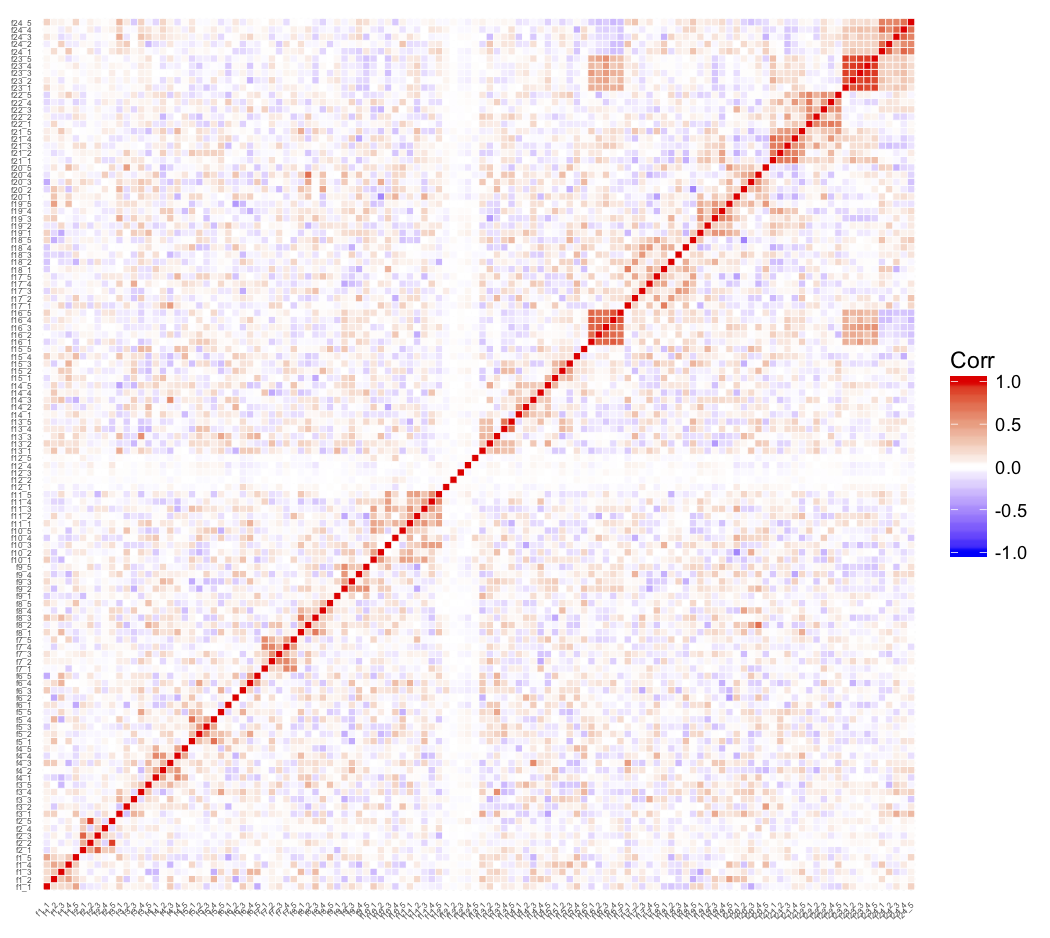}
    \caption{Pearson correlation between BBOB instances using the embeddings calculated with 38 singular values.}
    \label{fig:BBOB_embedded_space_38}
\end{figure}

The pairwise Pearson correlation between the embedded feature vectors of the BBOB instances are plotted in   
% using the new generated embeddings are presented in 
Figure~\ref{fig:BBOB_embedded_space_38}.
% Further, the proposed methodology was used to calculate an embedding for each benchmark problem instance. Since, we are interested in high accuracy (i.e., lower approximation error from the decomposition) and the computational efficiency is not a problem dealing with size of a matrix involved in our experiment, we used the 38 singular values to calculate the embeddings. The Pearson correlation between BBOB instances using the new generated embeddings are presented in Figure~\ref{fig:BBOB_embedded_space_38}.
%The new representation of the instances is in the space where different individual data dimensions are uncorrelated. For this reason, we used the SVM decomposition and decided to use all singular values (in our case 38), since we want to have higher accuracy. Also, we should mention that we are working with 38 landscape features, so the time is not a problem for our experiments and we want to have the best approximation from the decomposition. The Pearson correlation between BBOB instances using the new generated embeddings are presented in Figure~\ref{fig:BBOB_embedded_space_38}.
The diagonal pattern in this figure corresponds to a positive pairwise correlation of instances that belong to the same problem, in almost all cases.  
% shows that in most cases the same problem instances are positively correlated. 
High positive correlations are detected within instances of the following problems: $f_7$, $f_{16}$, $f_{21}$, $f_{22}$, $f_{23}$, and $f_{24}$. 
In addition, all instances from $f_{23}$ are positively correlated with all instances from $f_{24}$ and $f_{16}$. The third and fourth instances from the $f_{11}$ are more correlated than their correlation with the other three instances from the same problem. There is no obvious linear correlation between instances of a basic problem and its rotated variant (i.e., $f_{3}$ and $f_{15}$, and $f_{8}$ and $f_{9}$, respectively). Though many cases with missing strong linear correlation between different instances of the same problem might look strange at first, it is in fact what one should expect when considering that, by the way instances are generated in BBOB, the different instances are essentially forcing the sampling to focus on different parts of the same base problem -- a topic that we will return to in Figure~\ref{fig:corr_embed_invariant}. 
% The reason is in the sampling, which is done on different parts of problem search space for each problem instance. 

Comparing the correlations between the BBOB instances in the original and in the embedded space (Figures~\ref{fig:BBOB_feature_space_cor} and~\ref{fig:BBOB_embedded_space_38}), it is obvious that using the embeddings we see the positive correlations between the instances that belong to the same problem. We also detect 
% are able to distinguish between instances the same problem instances and also find 
groups of instances that are correlated with instances from other problems, which was not possible in the original feature space. These results can be further used in other applications. For example, working on automated algorithm design, it will be enough to take only one instance from the $f_{23}$ problem, since all of them are highly positive correlated, while for the $f_{11}$ problem all five instance should be involved, since there is only weak positive correlation between them. The correlation threshold for selecting the instances should be done by a subject-matter expert and depends on the application. %The application of the matrix factorization approach to such instance selection questions forms an important path for our future work. 
%Apart from the classification task studied in Section~\ref{sec:classificaiton}, we have to leave this questions for instance seclection for future work. 
%Comparing the results presented in figures~\ref{fig:BBOB_feature_space_cor} and~\ref{fig:BBOB_embedded_space_38}, we can conclude that by representing the BBOB instances using the new embeddings we are able to distinguish between instances from the same problem and also find the instances that are correlated with instances from other problems, which was not possible when we used them described in the original feature space. Additionally, the obtained results can be used in other applications. For example, if we are working on automated algorithm design or parameter tuning, then it will be enough to take only one instance from the $f_{23}$ problem, since all of them are highly positive correlated, while for the $f_{11}$ problem we can involve all five instance since there are weak positive correlation. Also, we should point that the correlation threshold for selecting the instances should be done by a subject-matter expert and depends on the application.
Another striking pattern in Figure~\ref{fig:BBOB_embedded_space_38} is the white cross %``+'' 
at the instances of problem $f_{12}$. These white tiles indicate that there is no linear correlation between %within 
the five instances of this problem, nor with any of the other $115$ considered BBOB instances.

\begin{figure*}[tb]
   \subfloat[Representation with 18 invariant features on translation and scaling in the original space.]{\label{fig:corr_embed_original}
      \includegraphics[width=.49\textwidth]{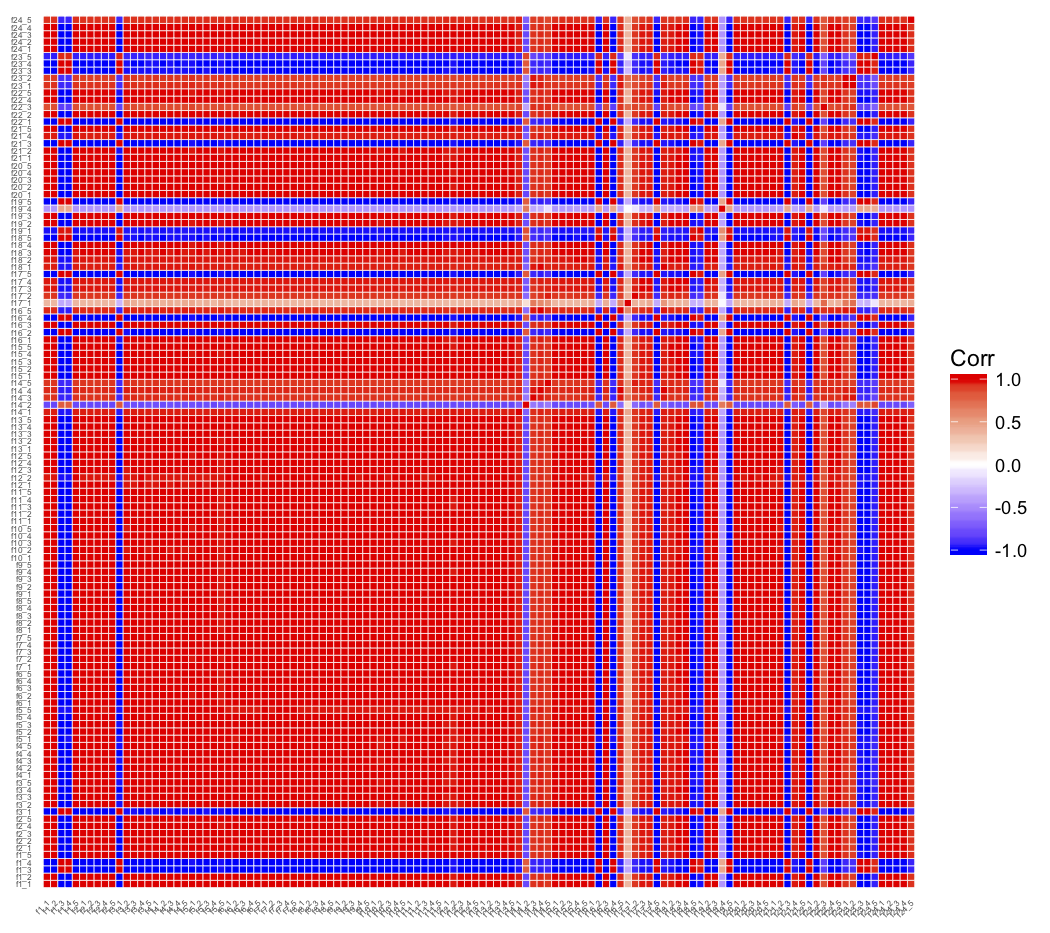}}
~
     \subfloat[Embeddings with 18  singular values for the features invariant on translations and scaling.]{\label{fig:corr_embed_invariant}
      \includegraphics[width=.49\textwidth]{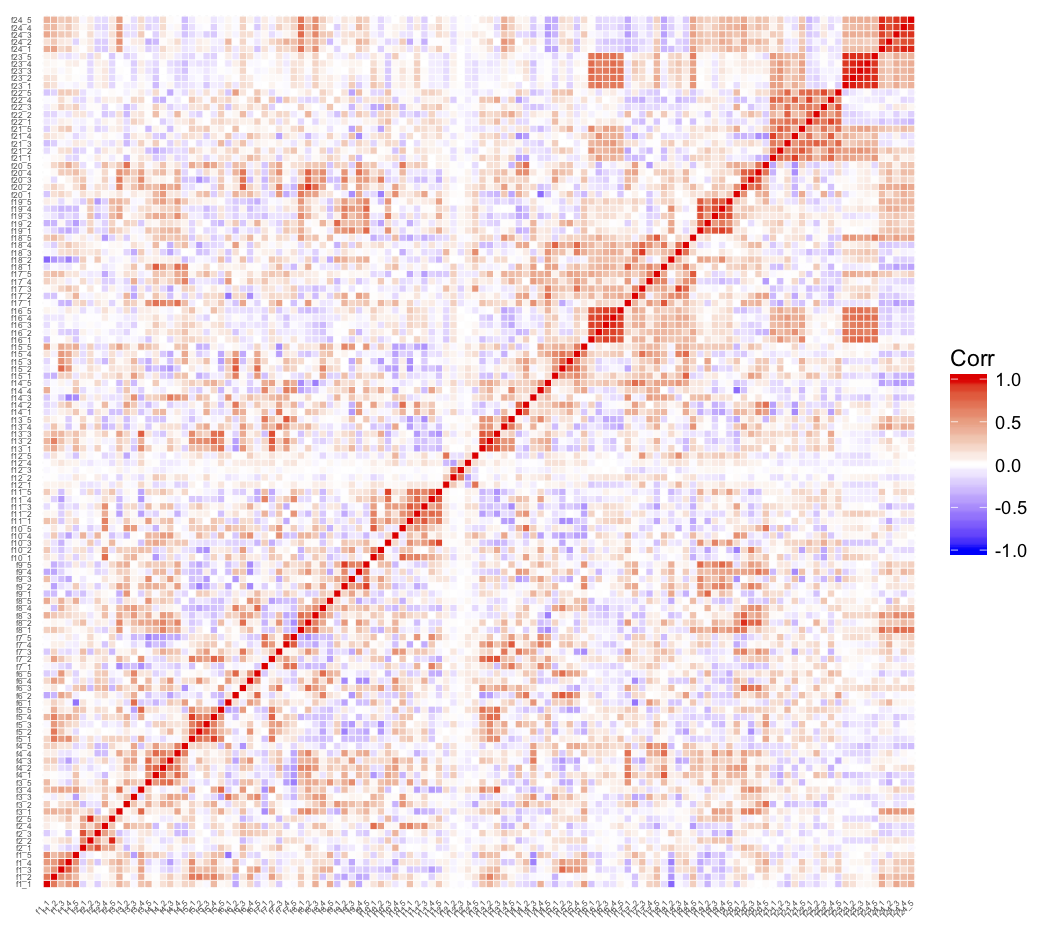}}
 \caption{Pearson correlation between BBOB instances.}\label{fig:BBOB_pearson_spearman}
\end{figure*}

In Figure~\ref{fig:corr_embed_invariant} we show the Pearson correlation between the embedded feature vectors that comprise only the 18 features that were classified in~\cite{vskvorc2020understanding} as invariant to translations and scaling.  
% Additionally, the Pearson correlations between the instances embeddings calculated with the 16 invariant features on translations and scaling reported in~\cite{vskvorc2020understanding} are presented in Figure~\ref{fig:corr_embed_invariant}. 
Comparing them to the correlations presented in Figure~\ref{fig:BBOB_embedded_space_38}, we observe the same patterns, but with different correlation intensities (we recall that the selection of a correlation threshold can be adapted with regard to the application that is being solved). Likewise, the correlation results in the original space described by the 18 selected features are presented in Figure~\ref{fig:corr_embed_original}. The only difference appears in the second instance of the $f_{14}$, which has high negative correlation with most of the BBOB instances. This does not show in the plots using all 38 features, where this instance shows weak negative correlation with most of the other BBOB instances.  This result shows that features that are not robust with respect to translation and scaling can have big impact on the comparison between problems in the original feature space. However, comparing the embedding spaces obtained with 38 features and 18 features, similar patterns appear.

%%%%%%%%%%%%%%%%%%%%%%%%%%%%%%%%%
%%%%%%%%%%%%%%%%%%%%%%%%%%%%%%%%%
\subsection{Complementarity of HappyCat and HGbat Functions}
\label{sec:new}

To illustrate how to test instance similarity of problems that are not part of the training set, we study the correlation of the BBOB functions with the HappyCat and the HGbat problems introduced in~\cite{beyer2012happycat}. 

The instances belonging to the HappyCat problem have a unique global minimum. According to~\cite{beyer2012happycat}, they are difficult to be optimized using state-of-the-art black-box optimizers such as the CMA-ES~\cite{hansen2003reducing}, particle swarm optimization~\cite{kennedy1995particle}, and differential evolution~\cite{price2013differential}. 
HGbat is similar to HappyCat, but differs in one term where a degree 8 polynomial instead of a degree 4 polynomial appears. For our illustration, we have select one random instance per each problem. To find the embeddings of these instances, we project their original feature value representation into the subspace that was learnt and described by the 120 BBOB instances with all 38 singular values, which describes our learnt embedded space. The Pearson correlation between the instances is presented in Figure~\ref{fig:BBOB_26,27}, where we refer to the HappyCat instance as $f_{25}$ and to the HGbat instance as $f_{26}$.
\begin{figure*}
   \centering
   \includegraphics[scale=0.35]{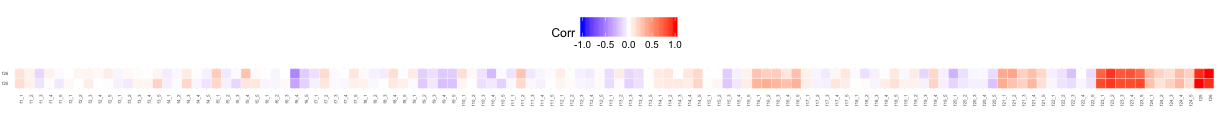}
   \caption{Pearson correlation between BBOB instances, HappyCat and HGbat problem using the embeddings generated using 38 singular values.}
   \label{fig:BBOB_26,27}
\end{figure*}
As expected, the HappyCat instance is highly positively correlated with the HGbat instance. Both of them are also highly positively correlated with all instances of $f_{23}$ (the so-called ``Katsurra function"). Additionally, both instances are positively correlate with all instance of $f_{16}$ (the ``Weierstrass function"),  with $f_{21}$ (Gallagher's Gaussian 101-me Peaks function), and $f_{24}$ (``Lunacek bi-Rastrigin function"). In~\cite{beyer2012happycat}, the authors discuss that the HappyCat function shares local similarities with the sharp ridge function, $f_{13}$. Looking in our results, however, the feature values do not seem to confirm (or be able to detect) such a correlation, raising the question if some essential properties are not captured by the selected features. These examples show that measuring instance correlation can be interesting also from a feature extraction point of view. 

\subsection{Sensitivity Analysis wrt Low-rank Approximations}
\label{sec:lowrank}

Our previous analyses were done using all 38 singular values, since we were interested in high approximation accuracy obtained from the SVD. To explore the sensitivity of the results, different low-rank matrix approximations were generated by selecting a smaller number of singular values (starting from 6 and going up to the full set of 38 singular values in steps of 1) from which the instance embeddings are calculated. The non-graphical solution to the Cattell subjective scree test estimator~\cite{raiche2013non} indicates that 16 singular values are %is 
a good starting point. To estimate the accuracy using the low-rank approximation $X_r$, the Frobenius norm $\vert\vert X-X_r\vert\vert_F$ of the difference between the original matrix $X$ and its low-rank approximation $X_r$ is calculated (Figure~\ref{fig:Frobenius_norm_1250}). The result shows that when the number of singular values increases, the approximation is much better, which is expected. 

\begin{figure}[!t]
 \centering
  \includegraphics[scale=0.28]{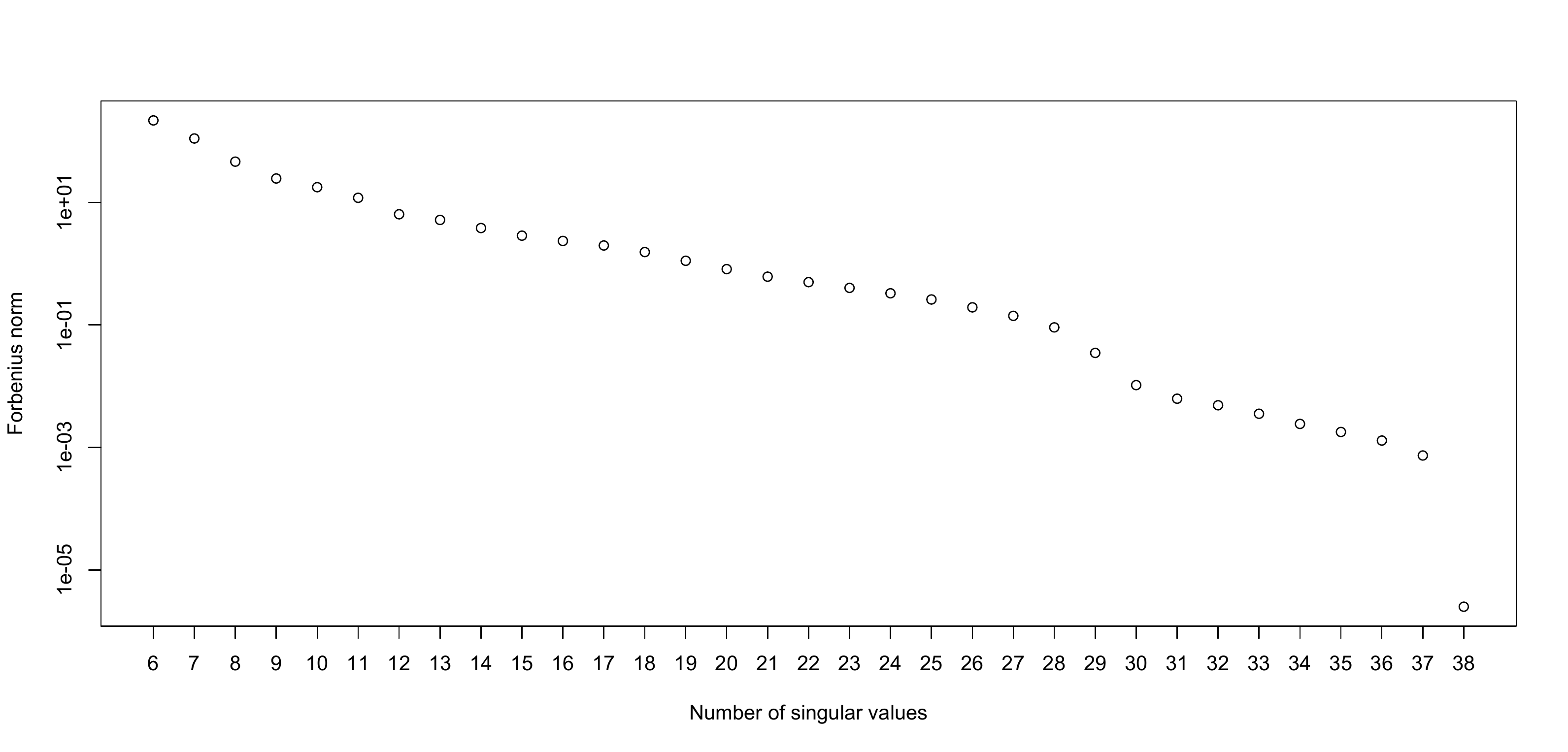}
   \caption{Frobenius norm of the difference between the original matrix $X$ and its low rank approximation $X_r$, for different numbers $r$ of singular values.}
  \label{fig:Frobenius_norm_1250}
\end{figure}
Figure~\ref{fig:BBOB_instances_sensitivity_analysis} presents the Pearson correlations between BBOB instances using the embeddings calculated with 19 and with 30 singular values, respectively. Comparing them, we observe once again similar patterns. By starting with the smaller number of singular values (Figure~\ref{fig:BBOB_19}) and going to a larger number of singular values (Figures~\ref{fig:BBOB_30} and~\ref{fig:BBOB_embedded_space_38}, respectively), we just obtain higher approximation accuracy. This indicates that in cases where computational efficiency is also important, a smaller number of singular values can be used. To do this, the Frobenius norm of the difference between the original and the low-rank approximation can be a good indicator for the trade-off between time and accuracy.

\begin{figure*}[!t]
 \subfloat[Embeddings calculated with 19 singular values.]{\label{fig:BBOB_19}
      \includegraphics[width=.49\textwidth]{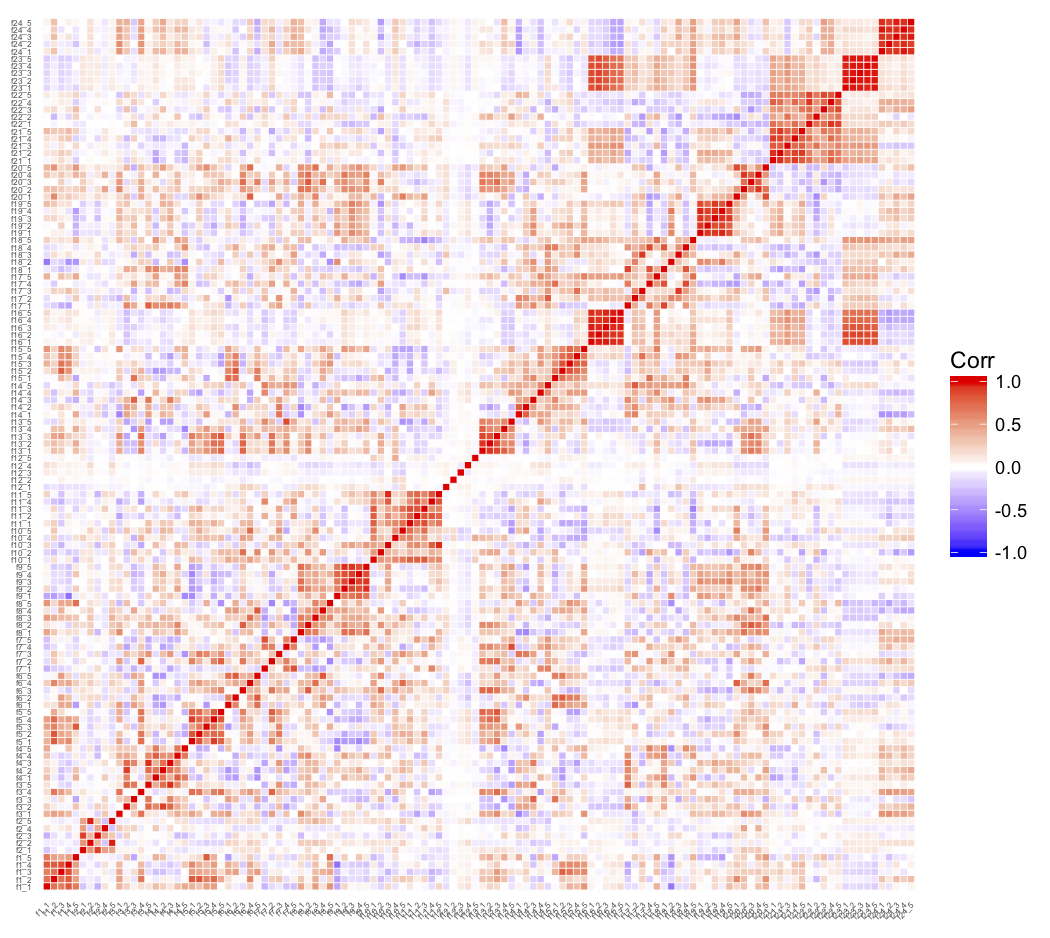}}
~
  \subfloat[Embeddings calculated with 30 singular values.]{\label{fig:BBOB_30}
      \includegraphics[width=.49\textwidth]{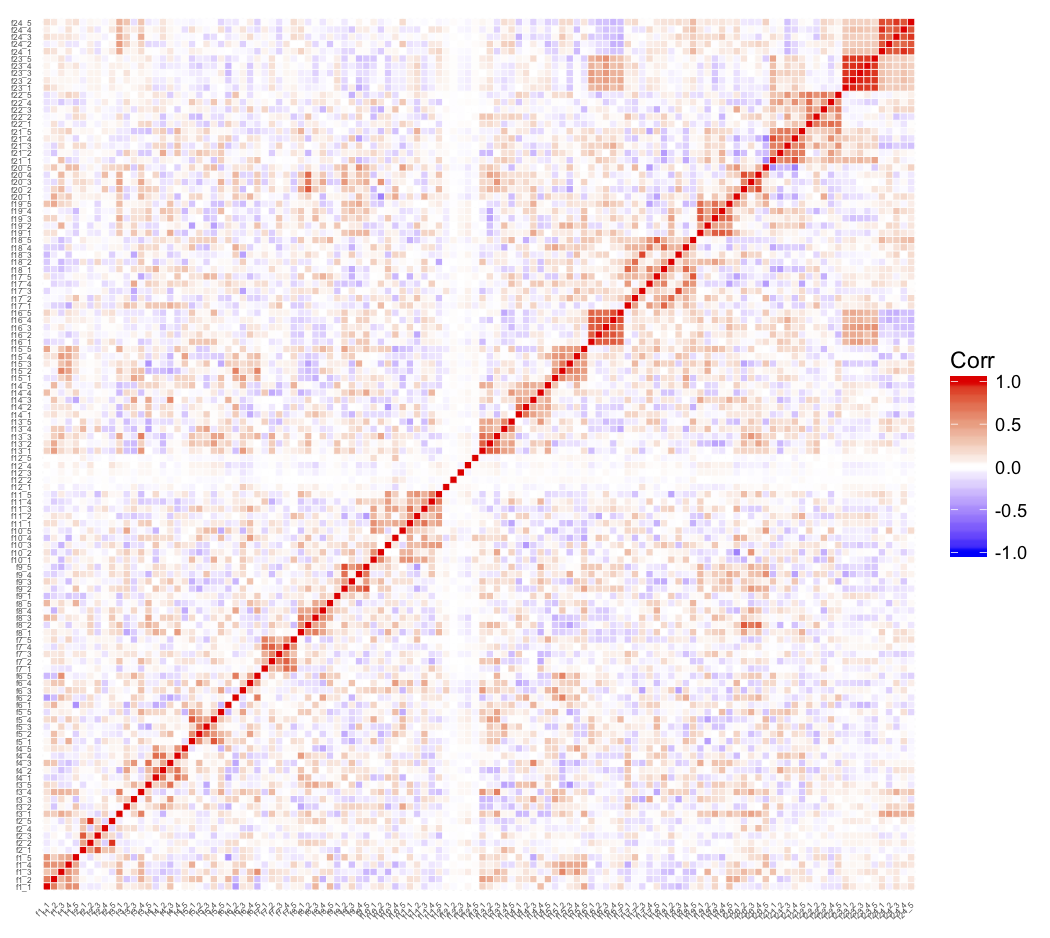}}

   \caption{Pearson correlation between BBOB instances using their embeddings. }\label{fig:BBOB_instances_sensitivity_analysis}
\end{figure*}

\subsection{Classification of BBOB instances}
\label{sec:classificaiton}

To further probe the utility of the learned embeddings, we also performed a multi-class classification, where for each instance we ask to predict the ID of the problem that it belongs to. That is, in our case we have 24 classes (one for each BBOB function), and for each class we have 5 instances. To evaluate the classification results, we used stratified 5-cross fold validation, where each fold consists of the first, second, third, fourth, and fifth instances for each problem, respectively. Then we repeated the process five times, where one fold was used for testing, and four of them for training the classifier. We should mention here that SVD is applied only on the training instances in order to learn their linear embedding, which are further used to train the classification model. In case of testing, the test instances were only projected into the subspace that was learnt and described by the training instances, which describes our learnt embedded space. 

For the classification, different state-of-the-art classifiers were trained exploring their hyperparameters by iterative grid search (see Table~\ref{tab:hyperparameters}). 
For the plots only the best-performing hyperparameter configuration for each classifier was selected. The classifiers that were used are: Stochastic Gradient Descent (SGD) (learning rate $\alpha$=0.01)~\cite{bottou2010large}, $k$-Nearest Neighbours ($k$=4)~\cite{liao2002use}, Support Vector Machine (SVM) with linear and radial basis kernel (regularization parameter $c=1.8$)~\cite{suykens1999least}, Random Forest (RF) with Gini index (number of trees 80) and entropy selection of features (number of trees 160)~\cite{liaw2002classification}, Extra Trees (ET) with Gini index (number of trees 80) and entropy selection (number of trees 200)~\cite{louppe2013understanding}, AdaBoost~\cite{hastie2009multi} and  Bagging~\cite{breiman1996bagging} with number of estimators 50 and Extra Trees as a base classifier with number of trees 10. The classification was performed using normalized (i.e., min-max scaling) and non-normalized features in both original and embedded space.

\begin{table}[tb]
    \centering
     \caption{Tested hyperparameters for the selected classifiers.}
     \resizebox{.49\textwidth}{!}{% <------ Don't forget this %
    \begin{tabular}{l|l}
    \hline
    Algorithm & Hyperparameters\\
    \hline
        SGD & $\alpha$ $\in$ \{10$^{-5}$, 10$^{-4}$, 10$^{-3}$, 10$^{-2}$, 10$^{-1}$,10$^{0}$\}  \\
        SVM & $c$ $\in$ \{0.2, 0.4, 0.6, 0.8, 1, 1.2, 1.4, 1.8, 2\}\\
        RF & $ntrees$ $\in$ \{5, 10, 20, 40, 60, 80, 100, 120, 140, 160, 180, 200\}\\
        ET & $ntrees$ $\in$ \{5, 10, 20, 40, 60, 80, 100, 120, 140, 160, 180, 200\}\\
        AdaBoost & 50 estimators, base classifier is ExtraTrees with 10 trees\\
        Bagging & 50 estimators, base classifier is ExtraTrees with 10 trees\\
        \hline
    \end{tabular}% <------ Don't forget this %
}
    \label{tab:hyperparameters}
\end{table}

The average accuracy obtained for the original feature space and for each of the embeddings calculated using 19, 27, 30, and 38 singular values, respectively, are summarized in Figure~\ref{fig:class_auc}. We also include in this figure the average classification accuracy when restricting the feature space to the 18 invariant features, once again in the original features space and in the embedded space (using all 18 singular values).
\begin{figure*}[!t]
 \subfloat[Non-normalized features.]{\label{fig:class_non-normalized}
      \includegraphics[width=.49\textwidth]{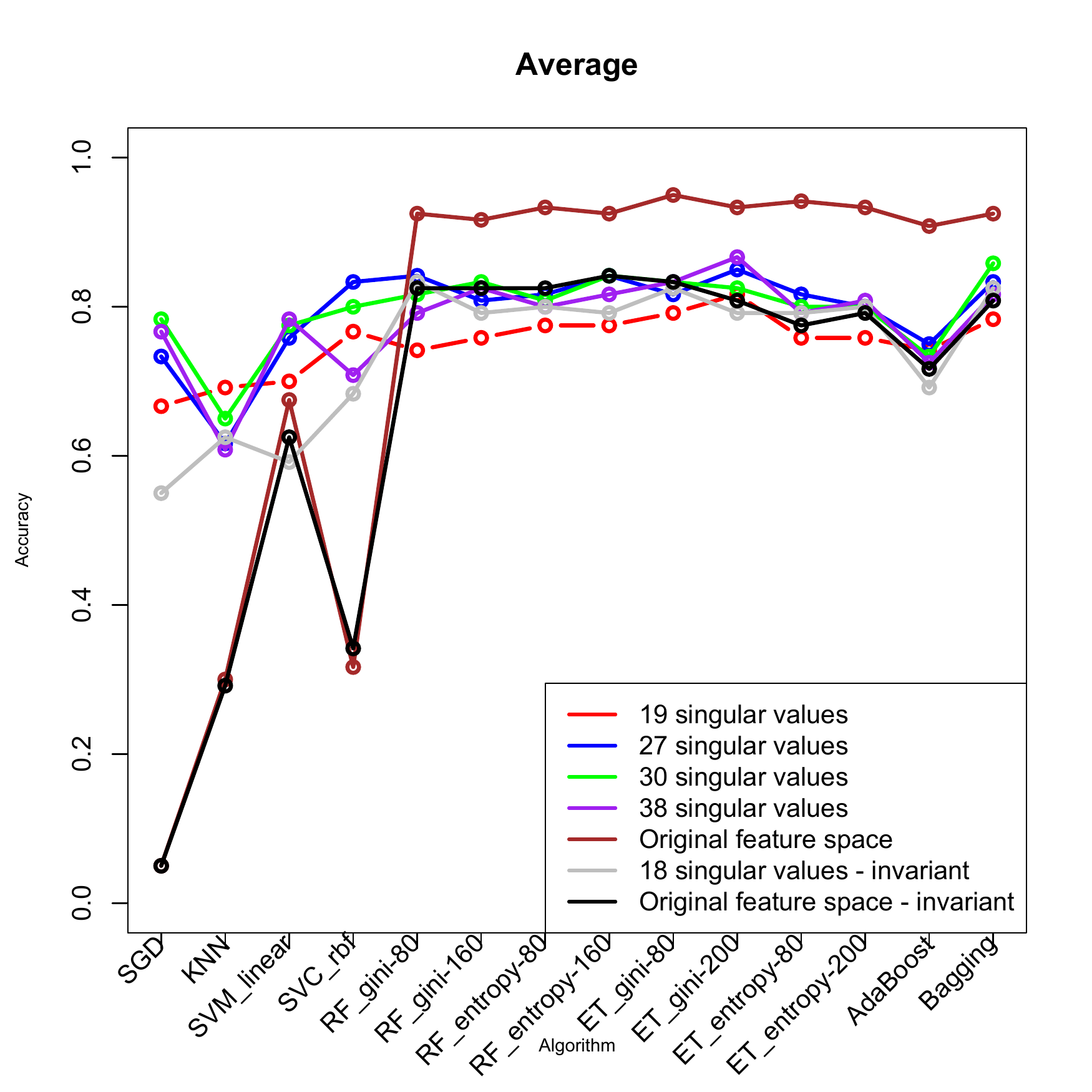}}
~
   \subfloat[Normalized features.]{\label{fig:class_normalized}
      \includegraphics[width=.49\textwidth]{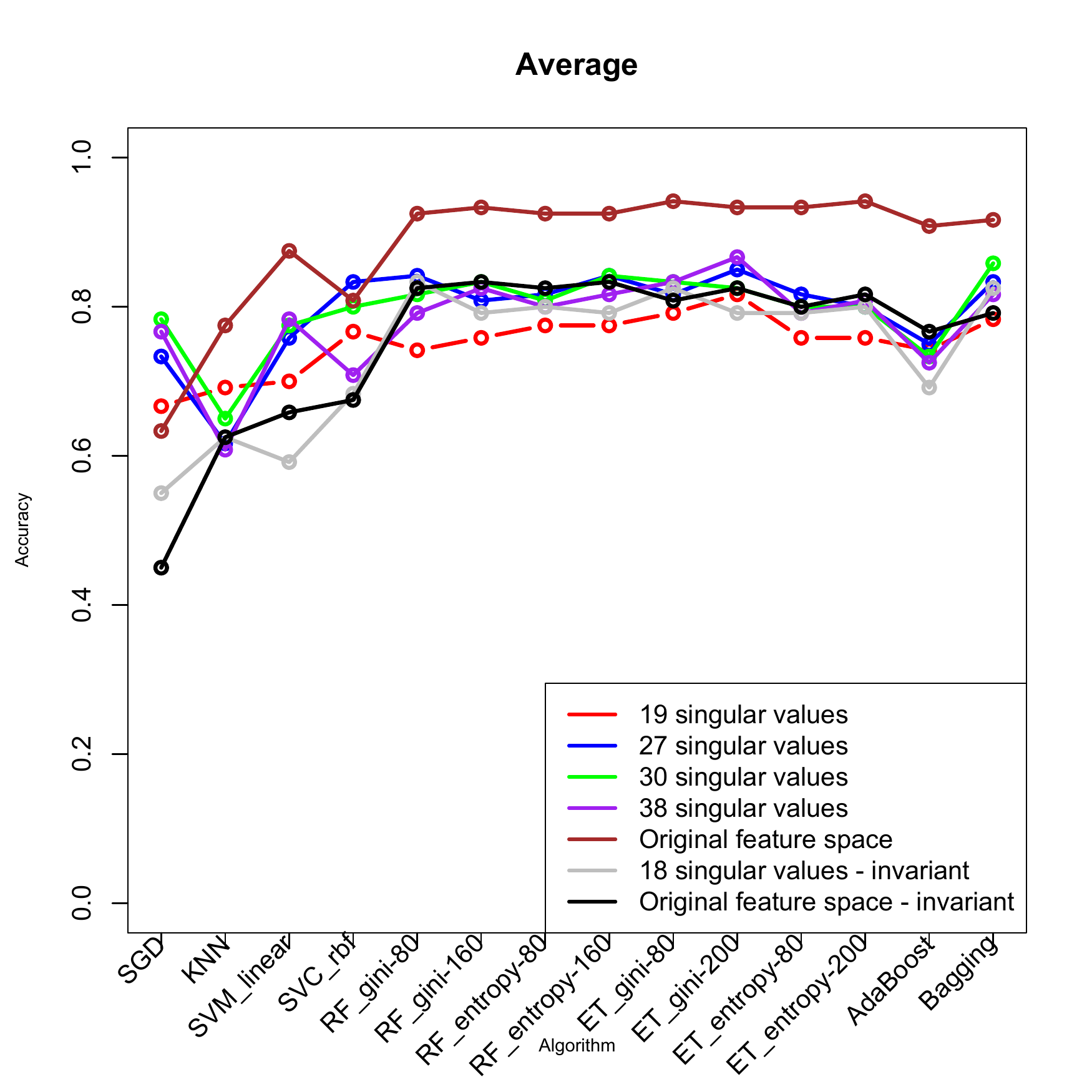}}
 \caption{5-cross fold validation accuracy for predicting the problem class. }\label{fig:class_auc}
\end{figure*}

Looking at the results plotted in Figure~\ref{fig:class_non-normalized} for the non-normalized features, the embeddings perform much better when they are used in a combination with classifiers that are not tree-based (i.e., SGD, KNN, and SVM), while the tree-based classifiers (i.e., RF, ET, AdaBoost, and Bagging) perform better with the original feature vectors. The absolute classification accuracy of the tree classifiers, however, is quite comparable (around 4-8\% percentile difference in accuracy). 
Comparing the results of the sensitivity analysis with respect to the number of singular values, we see that the embeddings obtained for 27, 30, and 38 singular values (blue, green, and purple lines, respectively) have higher accuracy than the embeddings obtained with 19 singular values. However, these differences are again quite small (around 3\% percentile). 
%Finally, comparing the classification results obtained in the original space using only the 16 invariant features with their embedded (full-rank) analogues, the same effects as observed for the case with all 38 features apply. 

We note that all embeddings that were generated using the data for all 38 features have better accuracy than both the original and the embedded feature vectors using the invariant features only. This shows that some additional features contained in the 38 contribute to the classification accuracy. These results open new directions for future research, where the proposed methodology can be explored and can help the process of feature selection. 
In the case of non-normalized features, the non tree-based classifiers never converge with their combination with the original features. In the case of normalized features, the original features space performs also well with the non tree-based classifiers, however, the absolute performance differences are again small. The classification results obtained with the embeddings are similar and are not affected by the normalization. The learned embeddings have already performed some kind of normalization on the data. 
%In the original space, however, some of the features have large data ranges, for which the normalization can help in the classification task. 

%Finally, we note that switching from mean to median does not change the overall structure of the charts in Figure~\ref{fig:class_auc}, but the dispersion of the accuracy is smaller for most of the classifiers trained with the embedded data, indicating a higher robustness for these classifiers than for those trained with original feature data.  
%accuracy plots look similar as the average accuracy plots and the accuracy dispersion is smaller in most of the classifiers trained with the embeddings, which points that they have better robustness than all classifiers trained with the original features.
%%%%%%%%%%%%%%%%%%%%%%%%%%%%%%%%%
%%%%%%%%%%%%%%%%%%%%%%%%%%%%%%%%%
\section{Conclusions and Future Work}
\label{sec:conclusions}

In this work we have investigated how representation learning %via matrix factorization 
can help avoid bias and redundancies in the feature sets commonly studied in the context of black-box optimization. More precisely, we have applied a matrix factorization approach to exploratory landscape analysis (ELA). Experimental results obtained for the BBOB benchmark set showed that it is difficult to distinguish between instances from the same benchmark problem when using the original feature vectors, whereas the new fingerprint representations achieved a much better fit. We have also trained classifiers to predict which problem a given instance belongs to. Results for the original feature sets and those trained with the embedded feature vectors provided similar results, demonstrating that the learned embeddings contain a lot of information from the original space. We have also analyzed the dimensionality reduction trade-off, showing that decent performance can be obtained also when reducing the dimension of the embedded vectors. 

The main motivation for our study is in deriving explainable ELA-based algorithm design techniques. The study indicates that representation learning can be a meaningful complement to computing-intensive feature construction and selection methods. Our study also opens a plethora of promising research directions, from which we summarize a few below. 
First and foremost, we will investigate the impact of our proposed approach on classical autoML tasks such as performance regression, algorithm selection, and algorithm configuration, both in the static~\cite{BelkhirDSS17,kerschke_automated_2019} as well as in the dynamic~\cite{Biedenkapp19,Vermetten19} case. 
We furthermore believe that the learned embeddings will prove useful for the selection as well as for the generation~\cite{Smith-MilesB15} of suitable benchmark instances. 

We have focused in this work on computing vector representation of benchmark problem instances through linear dimensionality reduction on the five-dimensional BBOB problems. An extension to higher dimensions is a direction that we plan to follow, to generate insight into how the BBOB problems in different dimensions relate to each other. Additionally, we plan to investigate the non-linear properties of the problem by applying non-linear dimensionality reduction techniques. 

In the longer term, we also plan on extending the analyses presented above to discrete and to mixed-integer optimization, where automated algorithm selection and configuration for general (i.e., not problem-specific) black-box optimization problems is still in its infancy~\cite{kerschke_automated_2019}.

\section*{Acknowledgment}

We thank Ale\u{s} Zamuda, Hao Wang, and Diederick Vermetten for several discussions on the content of this work. 

% \bibliographystyle{IEEEtran}
% \bibliography{IEEEfull}

\end{document}